\RequirePackage{amsmath}

\documentclass[runningheads]{llncs}
\usepackage[T1]{fontenc}
\usepackage{graphicx}
\usepackage{tcolorbox} 

\usepackage{amsfonts}
\usepackage{booktabs}
\usepackage{placeins}
\usepackage{bm}
\usepackage[misc]{ifsym} 
\usepackage[colorlinks=true]{hyperref}
\usepackage{amssymb}
\usepackage{caption}
\usepackage{graphicx,verbatim}
\usepackage{silence}
\usepackage{xcolor}  
\usepackage{tcolorbox}  
\newtcolorbox[auto counter, number within=section]{promptbox}[2][]{colframe=blue!70!black, colback=blue!5, coltitle=black, fonttitle=\bfseries, title=Prompt~\thetcbcounter: #2, #1}

\begin{document}
%




%

\author{Shaohao Rui \inst{1,2,4} \and
Haoyang Su \inst{2,3,4} \and
Jinyi Xiang \inst{1} \and \\
Lian-Ming Wu \inst{1} \and
Xiaosong Wang$^{\textrm{\Letter}}$\inst{2,4} 
}
\authorrunning{S. Rui et al.}

\institute{
\textsuperscript{1}Shanghai Jiao Tong University \\
\textsuperscript{2}Shanghai Innovation Institute \\
\textsuperscript{3}Fudan University, \textsuperscript{4}Shanghai AI Lab \\
\email{wangxiaosong@pjlab.org.cn} \\
\url{https://github.com/shaohao011/CardioCoT}
}


\title{CardioCoT: Hierarchical Reasoning for Multimodal Survival Analysis}
\titlerunning{Survival Analysis via VLM with Enhanced Reasoning}
    
\maketitle              

\begin{abstract}
Accurate prediction of major adverse cardiovascular events recurrence risk in acute myocardial infarction patients based on postoperative cardiac MRI and associated clinical notes is crucial for precision treatment and personalized intervention. Existing methods primarily focus on risk stratification capability while overlooking the need for intermediate robust reasoning and model interpretability in clinical practice. Moreover, end-to-end risk prediction using LLM/VLM faces significant challenges due to data limitations and modeling complexity. To bridge this gap, we propose CardioCoT, a novel two-stage hierarchical reasoning-enhanced survival analysis framework designed to enhance both model interpretability and predictive performance. In the first stage, we employ an evidence-augmented self-refinement mechanism to guide LLM/VLMs in generating robust hierarchical reasoning trajectories based on associated radiological findings. In the second stage, we integrate the reasoning trajectories with imaging data for risk model training and prediction. CardioCoT demonstrates superior performance in MACE recurrence risk prediction while providing interpretable reasoning processes, offering valuable insights for clinical decision-making.

\keywords{Risk Prediction \and Robust Reasoning \and Interpretability.}
\end{abstract}

\section{Introduction}
\begin{figure}
    \centering
    \includegraphics[width=1\linewidth]{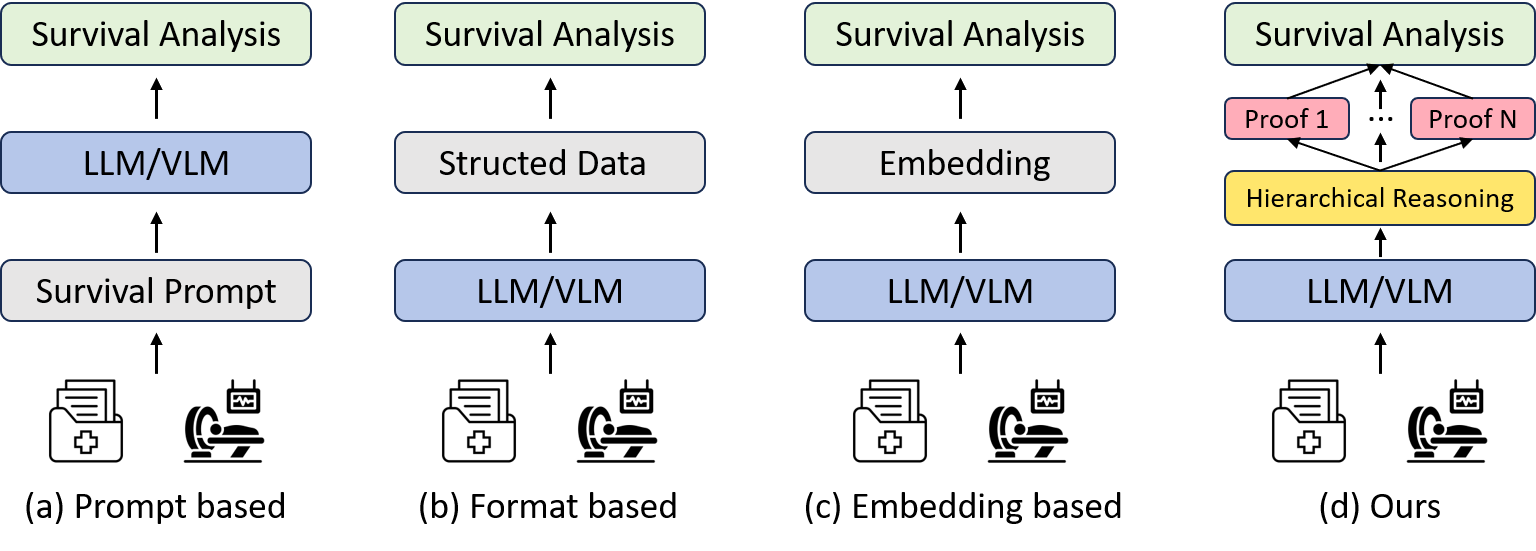}
    \caption{Comparison of LLM/LLM utilization in survival analysis: 1) Prompt-based: risk estimation via direct survival prompt; 2) Format-based: structured covariates extraction from unstructured data; 3) Embedding-based: semantic embedding extraction from pre-trained LLM/VLM; 4) Ours: hierarchical reasoning enabling evidential support and interpretable analysis.}
    \label{fig:definition}
\end{figure}

Survival analysis plays a pivotal role in accurately estimating the recurrence risk of Major Adverse Cardiovascular Events (MACE) for optimizing clinical treatment strategies and facilitating personalized interventions~\cite{wiegrebe2024deep,jeanselme2024review,jiang2024autosurv,jiang2024multimodal}. In recent years, a significant shift has been witnessed in survival analysis, with the rapid adoption of large language models (LLMs) and vision language models (VLMs) to extract valuable insights from unstructured medical data. This advancement extends beyond traditional methods that predominantly rely on structured covariates~\cite{jeanselme2024review,bush2017structured}, leading to enhanced predictive accuracy.

As shown in Fig.~\ref{fig:definition}, current methods using LLM/VLM for survival analysis can be classified into three types: 1) prompt-based methods directly query LLM/VLM to predict survival probabilities at specific time~\cite{han2023large}; 2) format-based methods leverage LLM/VLM's ability to extract structured data from unstructured clinical texts~\cite{agrawal2023large,truhn2024extracting,gero2023self,wei2023zero}, then facilitating the use of models like the Cox regression~\cite{cox1972regression}; 3) embedding-based methods~\cite{kim2024llm,kim2021deep,lee2021leveraging,kaka2022pretrained} involve using semantic embeddings derived from clinical texts to fine-tune specific survival model heads. In summary, prompt-based methods are effective for simple prediction tasks but struggle in more complex clinical scenarios. While format-based and embedding-based approaches have rapidly advanced in recent years, they primarily focus on risk stratification and often neglect the need for intermediate reasoning and model interpretability, both of which are crucial in clinical practice.

To address the growing clinical demand for interpretable reasoning in cardiovascular risk assessment, we propose \textbf{CardioCoT}, a novel two-stage hierarchical reasoning framework for survival analysis tailored to MACE recurrence risk prediction. Inspired by the careful consideration of multiple pieces of evidence in defining the endpoint (MACE recurrence) in clinical practice, CardioCoT first conducts hierarchical evidence-augmented reasoning through a self-refinement mechanism, generating multi-level predictions that integrate radiological diagnosis, complications, and MACE follow-up. These reasoning trajectories are subsequently used to tune LLM/VLMs, enabling systematic MACE-related clinical reasoning. In the second stage, CardioCoT aggregates multi-modal data from cardiac MRI scans and hierarchical model-generated textual reasoning processes for final survival analysis. Unlike conventional end-to-end approaches, CardioCoT explicitly models intermediate hierarchical clinical evidence pathways, achieving dual advancements in predictive accuracy and interpretability.

In summary, our contributions are three-fold: 1) We introduce CardioCoT, the first reasoning-enhanced survival analysis framework that emphasizes the need for reasoning pathways in clinical prediction; 2) We demonstrate an evidence-augmented self-refinement scheme to build a hierarchical set of reasoning trajectories for tuning our CardioCoT model for survival analysis; 3) Experiments on a real-world in-house dataset demonstrate a 7.53\% c-index improvement of the proposed model compared to the state-of-the-art method.
\begin{figure}[t]
    \centering
    \includegraphics[width=1\linewidth]{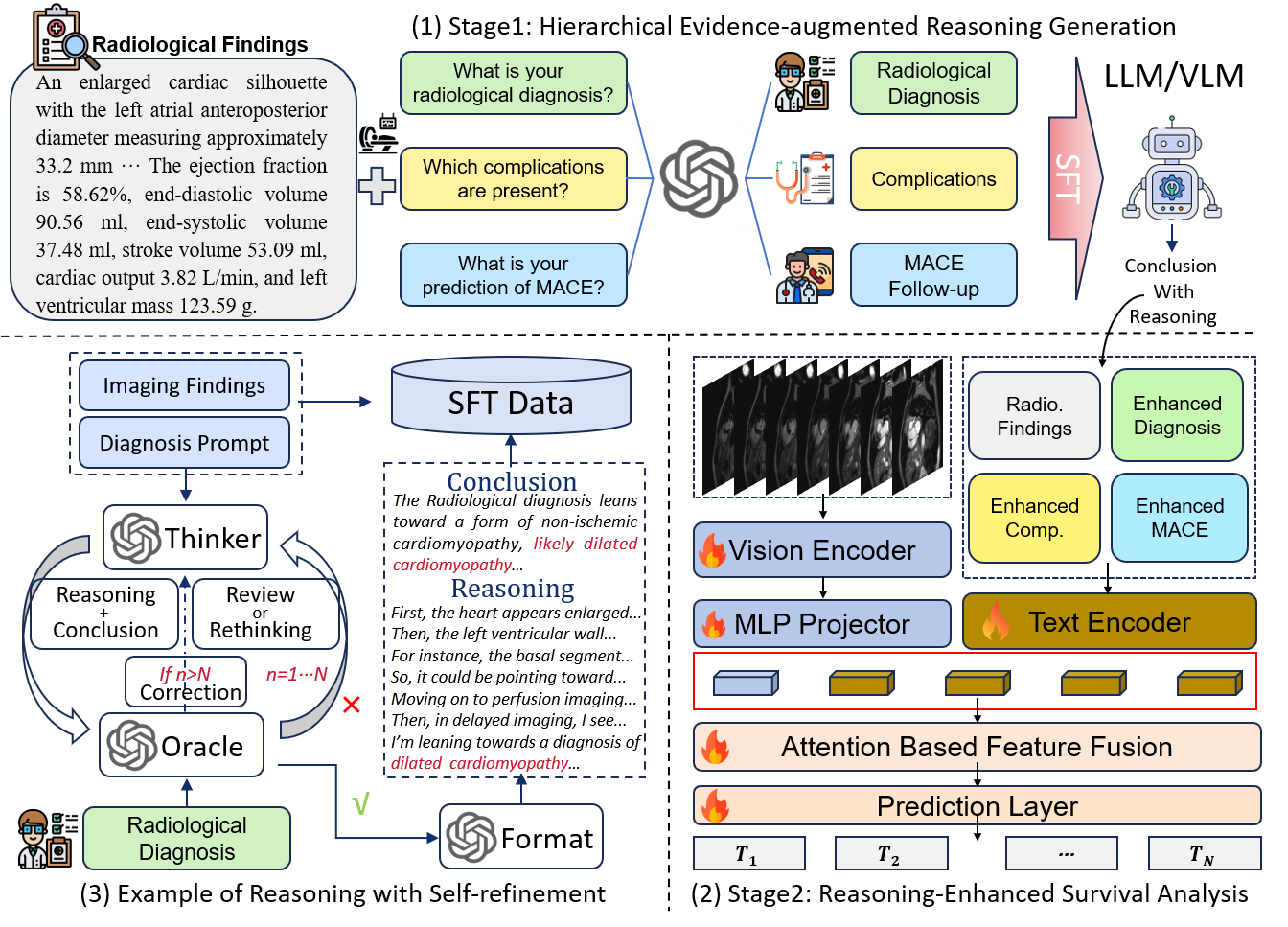}
    \caption{The proposed CardioCoT consists of two stages: (1) GPT-4o generates hierarchical evidence-augmented reasoning with self-refinement, fine-tuned on advanced LLMs/VLMs; (2) MRI scans and reasoning outputs are encoded by vision and text encoders, fused via an attention module for final risk prediction.}
    \label{fig:overview}
\end{figure}

\section{Method}
The proposed two-stage hierarchical reasoning-enhanced survival analysis framework is illustrated in Fig.~\ref{fig:overview}. In the first stage, MRI scans and radiological findings serve as inputs; during radiological diagnosis, complications and MACE follow-up act as endpoint-related evidence. A Chain-of-Thought (CoT) like prompt guides the Thinker Model in generating detailed reasoning and final conclusions through evidence-augmented self-refinement. The resulting $\langle \text{Inputs}, \text{Reasoning} \circ \text{Conclusion}\rangle$ pairs are then used to fine-tune VLMs/LLMs for clinical applications. In the second stage, image features are integrated with hierarchical textual features derived from reasoning-enhanced endpoint-related evidence, ensuring a more comprehensive recurrence risk prediction.

\subsection{Hierarchical Evidence-augmented Reasoning Generation}

Fig.~\ref{fig:overview}(3) illustrates the process of generating a self-refinement reasoning chain within hierarchical reasoning for radiological diagnosis. This process begins by querying the \textbf{Thinker} model (i.e., GPT-4o~\cite{achiam2023gpt}) with a diagnosis prompt \( {P}_d \), then generating an initial response \( \langle R_0, C_0 \rangle = \text{Thinker}(\mathcal{I}, P_d) \), where \( \mathcal{I} \) consists of MRI scans and corresponding radiological findings. Here, \( R_0 \) represents the initial reasoning and \( C_0 \) denotes the preliminary diagnostic conclusion. Following this, the \textbf{Oracle} model (i.e., GPT-4o~\cite{achiam2023gpt}) is employed to verify the consistency of the reasoning with the evidential radiological diagnosis \( E_{d} \). If the generated conclusion \( C_0 \) does not match \( E_{d} \), the reasoning undergoes iterative refinement through the \textbf{Review} or \textbf{Rethinking} processes, where both the reasoning and the conclusion are updated in each step.

This iterative process continues until either a correct diagnosis is achieved or the maximum number of iterations, \( N \) (i.e., 2), is reached. If \(C_{N}\) still does not match \( E_{d} \), an \textbf{Inference Correction} strategy is employed. In this phase, the correct diagnosis is explicitly provided to the Thinker model, instructing it to correct its reasoning and conclusion accordingly. The entire self-refinement process can be formally described as follows:
\begin{equation}
\langle R^*, C^* \rangle = 
\begin{cases} 
\text{Thinker}(\mathcal{I}, R_{n-1}, C_{n-1}, P_r), & \text{if } \mathcal{M}(C_n, E_{d}), \; n = 1, \dots, N  \\
\text{Thinker}(\mathcal{I}, R_{N}, C_{N}, E_{d}, P_c), & \text{if } \mathcal{M}(C_N, E_{d}) \text{ does not hold.}
\end{cases}
\end{equation}
where \( P_r \) represents the selected self-refinement strategy, while \( P_c \) denotes the inference correction prompt. Finally, the pair \( \langle \mathcal{I} \circ P_d, R^F \circ C^F \rangle \) is sent to the SFT dataset for training LLM/VLMs, where \( R^F \) and \( C^F \) are the formatted versions of the refined reasoning \( R^* \) and the corrected conclusion \( C^* \). The strategies mentioned above are as follows:
\begin{tcolorbox}[colframe=black!40!white, colback=white, title={}]
\small
    \textbf{Review} \textit{I have carefully examined the reasoning and concluded that your conclusion is incorrect. Please go back and review the earlier steps in the reasoning process, and build a revised conclusion.}
    
    \vspace{0.1cm}
    
    \textbf{Rethinking} \textit{I have carefully examined the reasoning and concluded that your conclusion is incorrect. Please construct a new reasoning process and build a new conclusion.}
    
    \vspace{0.1cm}
    
    \textbf{Inference Correction} \textit{The correct answer is \{···\}. You need to correct the previous reasoning process and conclusion to ensure the final answer is right.}
\end{tcolorbox}
Hierarchical reasoning builds upon the diagnostic reasoning framework described above by integrating relevant complications and MACE follow-up data one after another. Specifically, the complications encompass four major types of cardiac complications: Microcirculation Dysfunction, Intramyocardial Hemorrhage, Ventricular Thrombus, and Ventricular Aneurysm. Meanwhile, MACE follow-up tracks the recurrence status of MACE over a specified time period.

\subsection{Injecting Reasoning Ability to LLM/VLMs}
Building upon the hierarchical self-refinement reasoning processes generated by GPT-4o~\cite{achiam2023gpt}, we perform supervised fine-tuning (SFT) across multiple open-source models, encompassing both LLMs and VLMs, to enhance their clinical applicability. The VLMs include Qwen2-VL~\cite{Qwen2-VL} and InterVL2.5~\cite{chen2024expanding}, while the LLMs include LLama3.1~\cite{dubey2024llama} and HuatuoGPT-o1~\cite{chen2024huatuogpt}. VLMs are trained using both imaging and textual data, whereas LLMs are fine-tuned exclusively on textual data. Both types of models leverage LoRA~\cite{hu2022lora} for parameter-efficient training. Furthermore, model sizes ranging from 7B to 70B are explored to evaluate the impact of inference-time scaling on final MACE recurrence risk prediction. 

Finally, we generate reasoning-enhanced endpoint-related evidence $\mathcal{I}^D$, $\mathcal{I}^C$, and 
$\mathcal{I}^M$, which serve as part of multimodal inputs for final survival analysis.

\subsection{Reasoning-Enhanced Survival Analysis}

End-to-end probability-level survival analysis using LLM/VLMs is challenging due to the limited availability of data and the modeling complexity. To address this, we adopt a two-stage survival analysis approach, where the enhanced reasoning generated in the first stage is integrated with radiological data and fed into the second-stage survival analysis model for comprehensive interpretation.

\noindent{\textbf{Vision Encoder}}. Given the large inter-slice distance in 3D cardiac MRI scans, we use a 2D convolution-based model, specifically the 2D DenseNet121~\cite{huang2017densely}, as our vision encoder. For a given 3D MRI scan $X_{i}^{MRI} \in \mathbb{R}^{1\times N\times H\times W}$ with $N$ slices, we perform early fusion by treating the $N$ slices as distinct channels. Then, we apply 2D global average pooling to the extracted feature map from $X_{i}^{MRI}$, followed by an MLP projection to map the image features into the textual feature space, resulting in the final image embedding $Z_{i}^{img} \in \mathbb{R}^{1 \times 768}$.

\noindent{\textbf{Text Encoder.}} \noindent Considering the requirement of long text encoding ability, we use a pre-trained T5 model~\cite{raffel2020exploring}, which has been extended to process longer input token sequences, as our text encoder. This model is employed to tokenize and encode $\mathcal{I}^{F}$, $\mathcal{I}^{D}$, $\mathcal{I}^{C}$, and $\mathcal{I}^{M}$, where $\mathcal{I}^{F}$ denotes the textual descriptions of radiological findings. Consequently, we obtain the corresponding embedding vectors $Z_{i}^{f}$, $Z_{i}^{d}$, $Z_{i}^{c}$, and $Z_{i}^{ma} \in \mathbb{R}^{1\times768}$.

\noindent{\textbf{Attention-based Feature Fusion.}} Denote the bag of features mentioned above as \( \{Z_{i,m}^{bag}\}_{m=1}^{M} \), where \( M \) represents the number of modalities. To obtain the final aggregated features, we employ attention pooling as utilized in~\cite{ilse2018attention,chen2022pan}, which can be formally expressed as:
\small{
\begin{equation}
    Z_{i}^{bag} = \sum_{m=1}^{M}a_{i,m}Z_{i,m}^{bag},a_{i,m} = \frac{\exp\left\{ {w} \left( \tanh ({V} {{Z}_{i,m}^{bag}}^{\top}) \odot \text{sig} ({U} {Z}_{i,m}^{bag}) \right) \right\}}
    {\sum_{j=m}^{M} \exp\left\{ {w} \left( \tanh ({V} {{Z}_{i,m}^{bag}}^{\top}) \odot \text{sig} ({U} {{Z}_{i,m}^{bag}}) \right) \right\}}
\end{equation}
}where \( {w}, {V}, {U} \) are learnable parameters, and \( a_{i,m} \) denotes the attention score of the \( m \)-th modality, indicating its contribution to the final fused feature.
\section{Experiments} 
\subsection{Dataset}
We conduct validation experiments on our method using a real-world in-house dataset. This dataset includes postoperative MRI examinations and follow-up MACE records of 871 patients, among whom 104 were confirmed to have experienced recurrence during follow-up. 
Clinical data are obtained through a review of medical records, telephone interviews, and outpatient clinic visits. The primary endpoint is the occurrence of MACE, defined as a composite of cardiovascular death, hospitalization for heart failure, and reinfarction. To evaluate our algorithm, the dataset is randomly split into training, validation, and test sets in a 6:2:2 ratio. Survival time is divided into four intervals based on uncensored patients, as in~\cite{chen2022pan,qu2024multi}, and applied to the remaining patients.

\subsection{Evaluation Metrics and Compared Methods}
We evaluate our method using three metrics: the concordance index (C-index), survival area under the curve (Survival AUC), and Kaplan-Meier (KM) survival analysis with the log-rank test. The C-index measures the model’s ability to discriminate between different risk groups, with higher values indicating better risk stratification. Survival AUC, as proposed in~\cite{chen2022pan}, corrects for optimistic bias arising from censorship in model performance evaluation. Both the C-index and Survival AUC range from 0 to 1, with higher values reflecting better predictive accuracy. The KM curve visualizes survival distributions over time, while the log-rank test assesses the statistical significance of differences between groups. A smaller \textit{p}-value from the log-rank test indicates stronger stratification power, suggesting a clearer distinction between survival curves. 

Given the lack of associated methods for MACE recurrence risk prediction, we compare our algorithm with two current SOTA methods for multimodal survival analysis: PORPOISE~\cite{chen2022pan} and MMDB~\cite{qu2024multi}. We also report the results of single modal experiments in the baselines for a comprehensive evaluation. To ensure fairness in the comparison, we conduct experiments using the same data splits across all experiments.

\subsection{Implementation Details}
All experiments are conducted on two NVIDIA H100 GPUs. We use LLaMA-Factory~\cite{zheng2024llamafactory} framework for SFT using default configurations. For survival analysis, each MRI scan is sampled at 10 evenly spaced slice and resized to $256\times256$. Vision encoder is fully fine-tuned, while text encoder is trained using LoRA~\cite{hu2022lora}. The network is trained using AdamW, with a learning rate of 1e-4 and weight decay of 1e-5 for 20 epochs. We use the same survival loss used in~\cite{chen2022pan}, and \(\beta\) is set to 0.4 for fair comparison.

\begin{table}[h]
\caption{Experimental results on a real-world in-house dataset. The best results are highlighted in bold, while the second-best results are underlined.}
\centering
\begin{tabular}{llccc}
\toprule
\multicolumn{1}{c|}{Methods}  & \multicolumn{1}{c|}{Param Size}   & \multicolumn{1}{c|}{Modality}   & \multicolumn{1}{c|}{C-index $\uparrow$} & Survival AUC $\uparrow$\\ \midrule
\multicolumn{5}{c}{\textit{Single Modal}}                                                                     \\ \midrule
\multicolumn{1}{c|}{DenseNet121~\cite{huang2017densely}}  &  \multicolumn{1}{c|}{-}     & \multicolumn{1}{c|}{Image}      & \multicolumn{1}{c|}{0.6858}  & 0.7373       \\
\multicolumn{1}{c|}{T5-Base~\cite{raffel2020exploring}}  &\multicolumn{1}{c|}{-}       & \multicolumn{1}{c|}{Text}       & \multicolumn{1}{c|}{0.7075}  & 0.7379       \\ \midrule
\multicolumn{5}{c}{\textit{Multi Modal}}                                                                         \\ \midrule
\multicolumn{1}{c|}{PORPOISE~\cite{chen2022pan}} &\multicolumn{1}{c|}{-} & \multicolumn{1}{c|}{Image+Text} & \multicolumn{1}{c|}{0.7571}  & 0.7491       \\
\multicolumn{1}{c|}{MMDB~\cite{qu2024multi}} &\multicolumn{1}{c|}{-}  & \multicolumn{1}{c|}{Image+Text} & \multicolumn{1}{c|}{0.7589}  & 0.7545       \\ \midrule
\multicolumn{1}{c|}{CardioCoT-Qwen2-VL~\cite{Qwen2-VL}}  &\multicolumn{1}{c|}{7B} & \multicolumn{1}{c|}{Image+Text} & \multicolumn{1}{c|}{0.7927}  & 0.7601       \\
\multicolumn{1}{c|}{CardioCoT-InternVL2.5~\cite{chen2024expanding}}&\multicolumn{1}{c|}{8B} & \multicolumn{1}{c|}{Image+Text} & \multicolumn{1}{c|}{0.8008}  & 0.7646       \\
\multicolumn{1}{c|}{CardioCoT-LLama3.1~\cite{dubey2024llama}} &\multicolumn{1}{c|}{8B} & \multicolumn{1}{c|}{Image+Text} & \multicolumn{1}{c|}{0.8008}  & 0.7613       \\
\multicolumn{1}{c|}{CardioCoT-LLama3.1~\cite{dubey2024llama}} &\multicolumn{1}{c|}{70B} & \multicolumn{1}{c|}{Image+Text} & \multicolumn{1}{c|}{\underline{0.8115}}  & \underline{0.7651}       \\
\multicolumn{1}{c|}{CardioCoT-HuatuoGPT-o1~\cite{chen2024huatuogpt}} &\multicolumn{1}{c|}{7B} & \multicolumn{1}{c|}{Image+Text} & \multicolumn{1}{c|}{0.7925}  & 0.7611      \\ 
\multicolumn{1}{c|}{CardioCoT-HuatuoGPT-o1~\cite{chen2024huatuogpt}} &\multicolumn{1}{c|}{70B} & \multicolumn{1}{c|}{Image+Text} & \multicolumn{1}{c|}{\textbf{0.8342}}  & \textbf{0.7712}       \\ \bottomrule
\end{tabular}
\label{tab:performance_comparison}
\end{table}
\begin{figure}[t]
    \centering
    \includegraphics[width=1\linewidth]{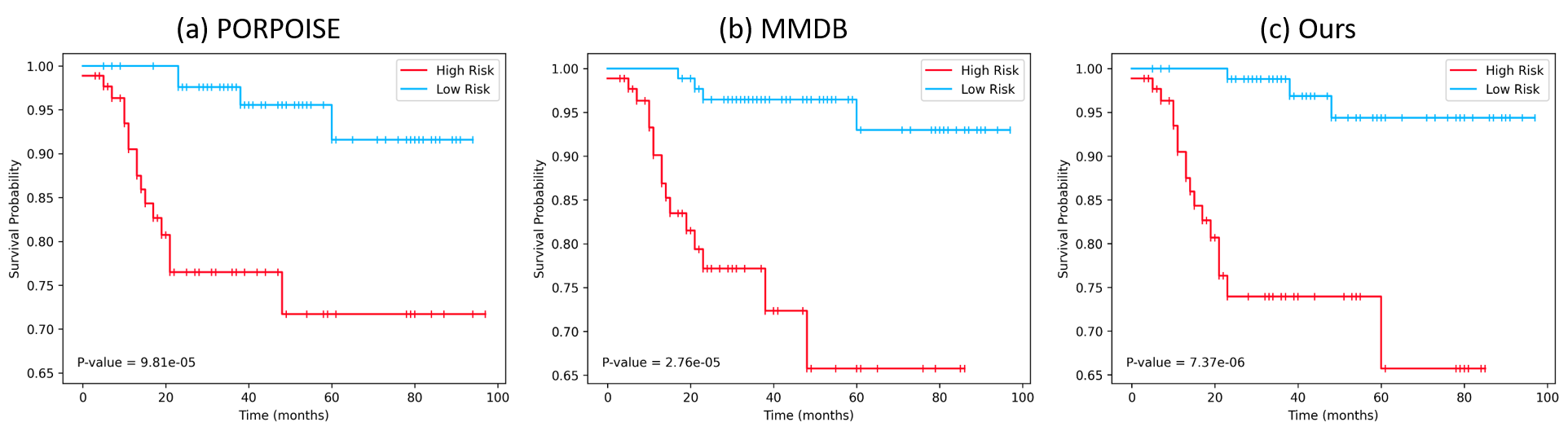}
    \caption{The KM analysis curves results.}
    \label{fig:km-curve}
\end{figure}

\subsection{Results}
As shown in Table~\ref{tab:performance_comparison}, 
a 7.53\% c-index improvement is achieved by the proposed model compared to the state-of-the-art method and single modal baselines. We also notice that the prediction performance gain is consistent with the increase in model size and domain fine-tuning can further benefit the final performance as well. 

Fig.~\ref{fig:km-curve} presents the Kaplan-Meier (KM) curves comparisons between our method and the state-of-the-art methods. The division between high-risk and low-risk groups is determined by the median predicted risk. A larger gap between the predicted survival probability curves for these groups indicates better risk stratification capability. Our method achieves a \textit{p}-value of 7.37e-6, which significantly outperforms the baseline methods, demonstrating its strong effectiveness in risk stratification.
\begin{figure}[!ht]
\centering
\begin{minipage}{0.48\textwidth}  
\captionof{table}{Ablation with reasoning-enhanced diagnosis (D.), complications (C.), and MACE follow-up (M.).}  
\centering
\begin{tabular}{ccc|cc}
\toprule
\multicolumn{3}{c|}{CoT} &  \multicolumn{2}{c}{Metrics} \\ 
\midrule
D. & C. & M. & C-index $\uparrow$ & Survival AUC $\uparrow$ \\ 
\midrule
\checkmark & & & 0.7692 & 0.7519 \\ 
& \checkmark & & 0.7704 & 0.7525 \\ 
& & \checkmark & 0.7737 & 0.7524 \\ 
\checkmark & \checkmark & & 0.7881 & 0.7563 \\ 
\checkmark & & \checkmark & 0.7787 & 0.7554 \\ 
& \checkmark & \checkmark & 0.7852 & 0.7453 \\ 
\checkmark & \checkmark & \checkmark & \textbf{0.8008} & \textbf{0.7613} \\ 
\bottomrule
\end{tabular}
\label{tab:cot_components}  
\end{minipage}
\hfill
\begin{minipage}{0.51\textwidth}  
\centering
\includegraphics[width=1.0\linewidth]{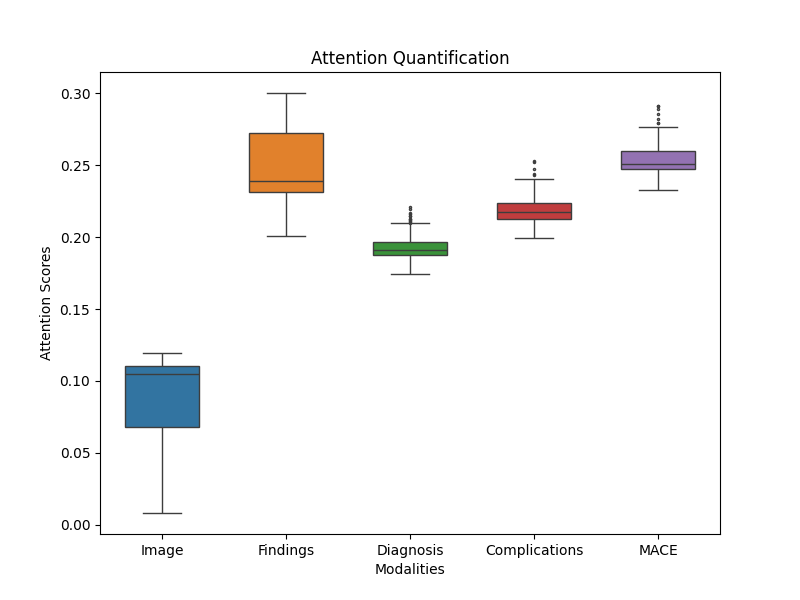}
\captionof{figure}{Attention quantification of different parts for survival analysis.}
\label{fig:attention-quanti}
\end{minipage}
\end{figure}

\begin{figure}[t]
    \centering
    \includegraphics[width=1\linewidth]{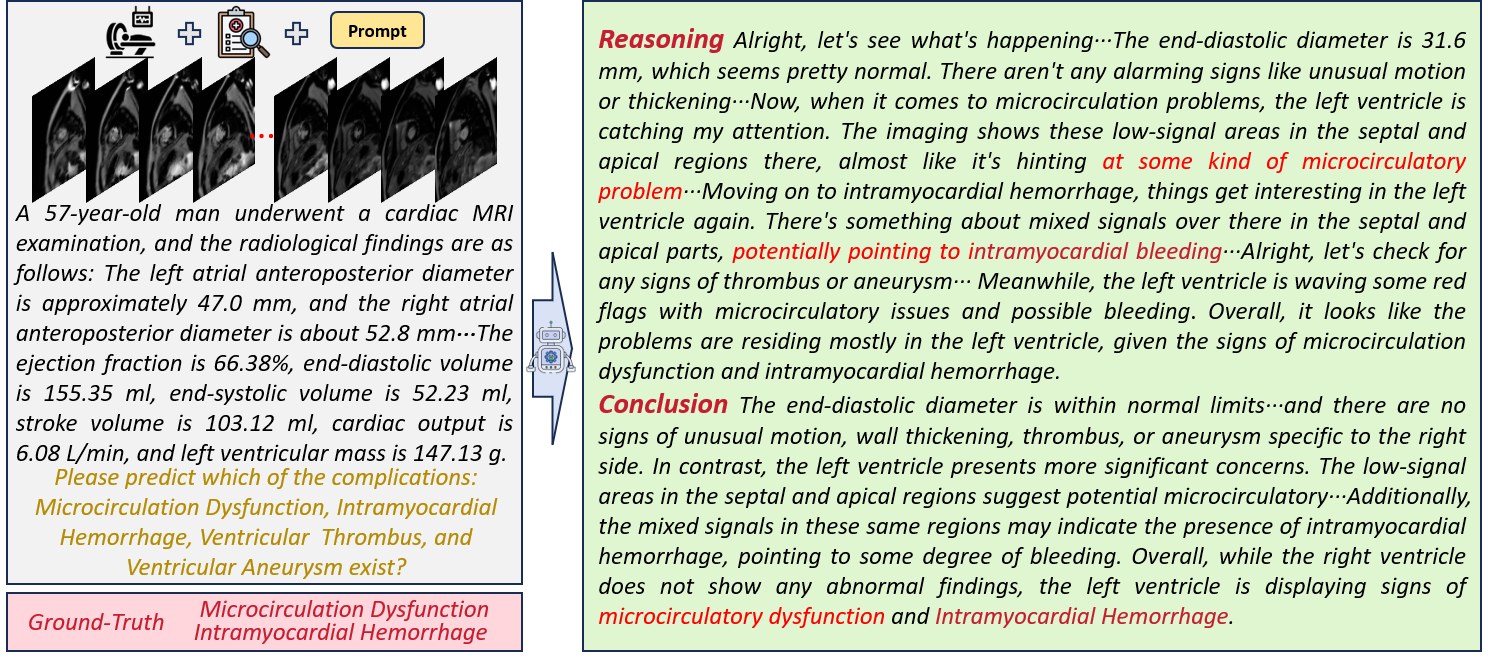}
    \caption{Qualitative result of a case for prediction complications with reasoning.}
    \label{fig:case_study}
\end{figure}
Additionally, given the introduction of hierarchical intermediate reasoning processes, our method offers better clinical interpretability. Attention score distribution in Fig.~\ref{fig:attention-quanti} clearly demonstrates the consistent effectiveness of our additional reasoning modules across different samples. Interestingly, the attention scores for imaging features exhibit greater variance while accounting for a smaller proportion, suggesting that radiological findings are significantly complemented by the added reasoning processes, which capture information that is difficult to extract directly from images. Fig.~\ref{fig:case_study} illustrates the enhanced complications prediction for a patient with MACE recurrence during follow-up. CardioCoT clearly demonstrates the intermediate reasoning process, providing a retrospective reference for clinical decision-making.

\noindent{\textbf{Ablation Study}}: We conduct a comprehensive ablation study on the three hierarchical reasoning-enhanced components proposed in our method. As shown in Table~\ref{tab:cot_components}, the experimental results highlight the critical role of each module and demonstrate that integrating all three reasoning components effectively improves overall performance.

\section{Conclusion}
In this paper, we propose CardioCoT, a two-stage hierarchical reasoning-enhanced survival analysis framework for MACE recurrence risk prediction. By integrating multi-level reasoning and using advanced LLM/VLMs, our method improves both interpretability and predictive accuracy, achieving state-of-the-art performance. Experimental results validate its effectiveness in risk stratification and clinical decision support. In the future, we will further investigate and validate the applicability of this reasoning enhancement method across additional clinical scenarios.



%
%
%
\bibliographystyle{splncs04}
\bibliography{reference}
\end{document}